%% file: uai2026.tex
\documentclass[accepted]{uai2026} 

\usepackage[british]{babel}

\usepackage{amsmath}
\usepackage{amssymb}
\usepackage{mathtools}
\usepackage{amsthm}
\usepackage{nicefrac}

\usepackage{graphicx}
\usepackage{booktabs}
\usepackage{multirow}
\usepackage{tabularx}
\usepackage{array}
\usepackage{longtable}

\usepackage{enumitem}
\usepackage{setspace}
\usepackage{wrapfig}
\usepackage{lscape}
\usepackage{cancel}

\usepackage{xcolor}
\usepackage[most]{tcolorbox}
\usepackage{caption}
\usepackage{subcaption}
\usepackage[size=scriptsize]{todonotes}
\usepackage{natbib}

\definecolor{myblue}{RGB}{0, 76, 153}
\definecolor{darkred}{RGB}{153, 0, 0}

\usepackage{hyperref}
\hypersetup{
  colorlinks=true,
  linkcolor={red!50!black},
  citecolor=myblue,
  urlcolor=black
}

\usepackage{tikz}
\usetikzlibrary{
  bayesnet,
  positioning,
  shapes.geometric,
  shapes.multipart,
  fit,
  calc,
  arrows.meta,
  bending,
  decorations.pathreplacing,
  angles,
  quotes,
  calligraphy
}

\let\olddiv\div
\usepackage{physics}
\let\div\olddiv

\input{math_notations.tex}

\newcolumntype{L}[1]{>{\raggedright\arraybackslash}p{#1}}

\theoremstyle{plain}

\newtheorem{example}{Example}
\newtheorem{remark}{Remark}
\newtheorem{definition}{Definition}

\theoremstyle{remark}

\newcommand{\nodeSet}{\ensuremath{\mathbf{V}}}

\allowdisplaybreaks

\title{A Unifying Perspective on Causal World Models: \\From Observations to Representations to Structure}

\author[1]{Avinash Kori}
\author[1]{Fabrizio Russo}

\affil[1]{
  Department of Computing, Imperial College London, London, UK\\
  
  \{a.kori21,fabrizio\}@imperial.ac.uk
}

\begin{document}

\maketitle

\begin{abstract}
    \vspace{-0.5cm}
World Models (WM) are increasingly seen as a foundation for intelligent agents that can predict, plan, and act beyond their training distribution.
In this paper, we study WMs from a causal perspective across multiple levels of abstraction, ranging from perceptual observations to building a conceptual representation of the structure governing the environment dynamics.
We argue that useful WMs must go beyond 
generative capabilities alone: they should also capture entity properties, entity-to-entity interactions, and entity-to-environment interactions that determine and explain the dynamics of a system.
We provide a formal definition of Causal WMs (CWMs) grounded in the tasks they are intended to support, connecting world modelling with existing work in causal representation learning, object-centric learning, causal discovery, structural causal models, and model-based decision-making.
Finally, we relate CWMs to the literature on identifiability, clarifying when the components of a WM can be recovered from data and up to which equivalence.
With this, we ground WMs in representations and structures that support causal reasoning and informed decision-making. 
\end{abstract}
\vspace{-0.3cm}
\section{Introduction}\label{sec:intro}
\vspace{-0.3cm}
World models have become one of the central aspirations of modern Artificial Intelligence. 
They are invoked to describe agents that imagine, plan, and learn to act safely outside their training distribution~\citep{ha2018world,lecun2022path,matsuo2022deep,taniguchi2023world}.
The challenge is to \emph{build} such models from observations, trajectories, interventions, and domain knowledge, while avoiding two unhelpful extremes: treating WMs as an unstructured black box with no explicit variables or mechanisms, or assuming in advance that the right variables and mechanisms have already been specified, so that they can be directly used for causal intervention.
\vspace{-0.15cm}

Recent work occupies different points along this spectrum.
Predictive world models such as Dreamer learn compact latent states from sensory streams and use imagined rollouts for control across diverse tasks, but the learned states need not have an explicit causal or semantic interpretation~\citep{hafner2025dreamer}.
Hierarchical variants such as ResDreamer move closer to task-level reasoning by emphasising decision-relevant signals over photorealistic reconstruction, while still treating the internal representation largely as a learned latent substrate~\citep{xu2026resdreamer}.
Joint-embedding approaches such as LeJEPA address a different part of the problem by connecting predictive representation learning with formal identifiability guarantees, but they do not by themselves specify the causal variables, mechanisms, or intervention targets needed for acting in the world~\citep{klindt2026lejeppa}.
Causal dynamics methods illustrate the complementary structural route: once state variables are given, sparse causal dependencies can improve generalisation and yield reusable state abstractions for downstream tasks~\citep{wang2022causal_dynamics, wang2024causal_bisimulation}.
Recent work on robust and general agents makes the motivation especially direct: agents need WMs that support causal reasoning about actions, mediation, and distribution shifts, not merely next-state prediction~\citep{richens2024robust}. However, even in this literature, WMs are often still operationalised primarily as transition models~\citep{richens2025general,ceriscioli2025agents}.

\begin{figure*}[h!t]
    \centering
    \begin{tikzpicture}
        \node[anchor=south west,inner sep=0] (ladder) at (0,0)
        {\includegraphics[width=\linewidth]{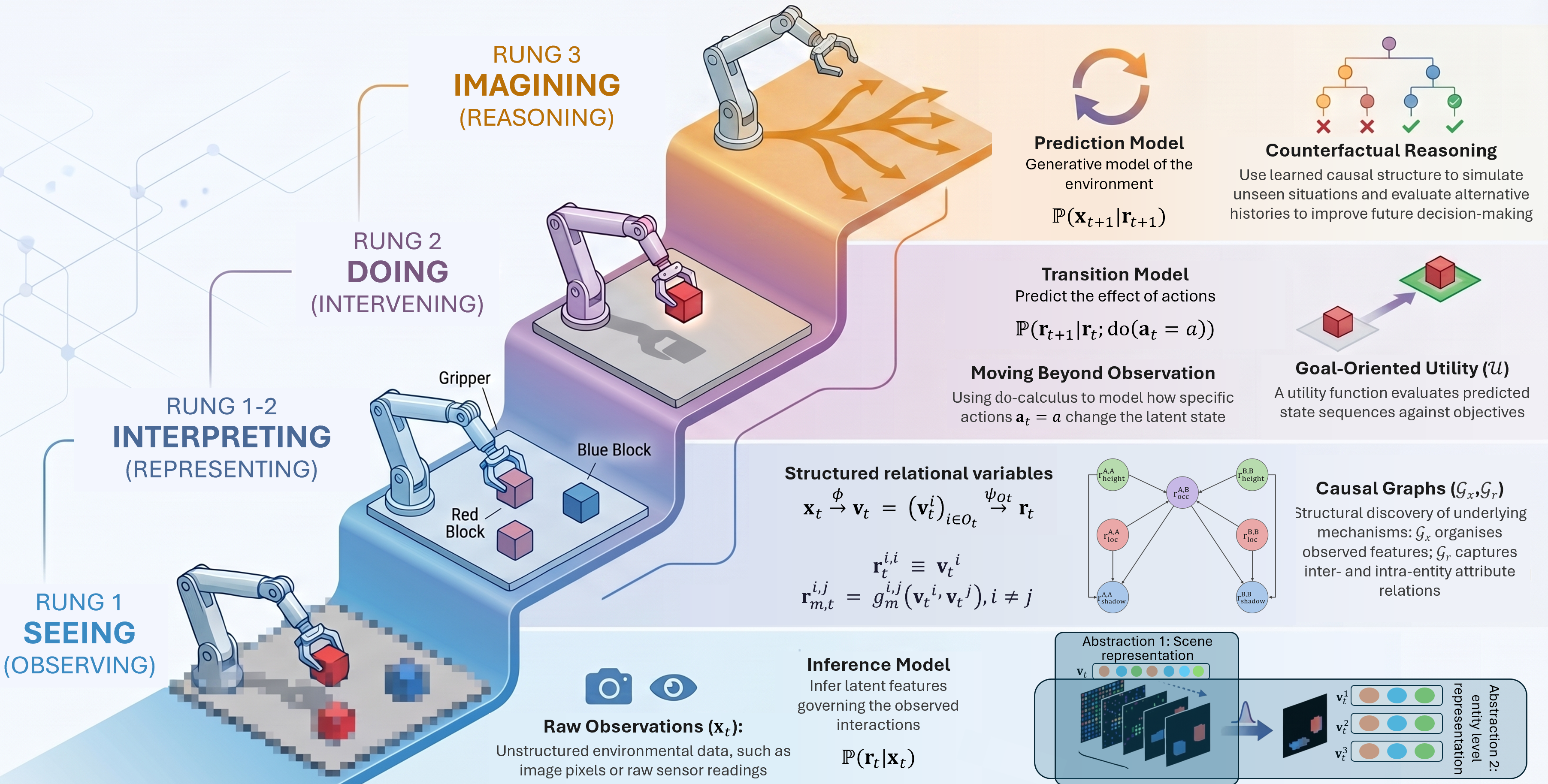}};
    \end{tikzpicture}
    \caption{Conceptual overview of a Causal Ladder of CWM: illustrated as a four-level progression from perception to intervention and imagination. Rungs 1, 2 and 3 follow Pearl's causal hierarchy~\citep{pearl2018bookofwhy,bareinboim2022on}. At the base, Rung 1 represents raw sensory perception, where raw observations $\rvx_t$ are processed by an inference model to infer entity-level features $\rvv_t$. In Rung 1–2, these features are lifted into the structured relational state $\rvr_t$, whose diagonal blocks retain entity attributes and whose off-diagonal blocks encode interactions. The graph $\gG_r$ represents causal mechanisms over this state; $\gG_x$ is used only when observations have themselves been decomposed into explicit variables. At Rung 2, agents can manipulate entities to maximise their utility. At the top, Rung 3 represents counterfactual and generative reasoning, enabling agents to analyse different possible scenarios.\\[-0.9cm]
    }
    \label{fig:CWM}
\end{figure*}
\vspace{-0.18cm}
\emph{Yet the phrase World Model often names an ambition more than a well-specified object.} 
Recurrent latent dynamics models, next-observation predictors, simulators, joint-embedding predictors, learned state abstractions, as well as the implicit knowledge stored in a large generative model have all been referred to as WMs~\citep{guan2023leveraging,ding2025understanding}. 
The difficulty is that this common label conflates distinct modelling commitments. 
A predictor specifies what is likely to happen next; a representation compresses observations by mapping \textit{unstructured} observations into usable structured variables representing relevant aspects of the world; and a causal model specifies mechanisms enabling interventional analysis.
What remains underspecified is how to connect these pieces towards building robust WMs with known properties and guarantees: which variables should be represented, whether they are identifiable from data, which relations among them are causal, when an action-conditioned prediction can be ascribed an interventional-effect interpretation, and when additional information or domain knowledge is required.

\noindent The problem we focus on is therefore how to turn observations into a CWM by combining three strands of work: causal representation learning~\citep{scholkopf2021toward, bengio2013generalized}, addressing when latent variables can be recovered from unstructured observations; causal discovery~\citep{ZANGA2022survey}, studying when cause-effect relationships over the structured variables (represented by $\gG_r$; see Figure~\ref{fig:CWM}) can be recovered; and model-based decision-making~\citep{ge2026review}, covering when actions and utilities can be used for controlled behaviour. When observations are explicitly decomposed into variables, $\gG_x$ denotes a graph over those observed or encoded variables. We do not assume $\gG_x=\gG_r$: their correspondence requires an explicit conditional-independence-preserving representation map. Faithfulness relates each distribution to its own graph and, together with the Markov and sufficiency assumptions, yields the usual Markov-equivalence limit for observational discovery over $\gG_r$~\citep{spirtes2000causation,verma1990equivalence}.
Figure~\ref{fig:CWM} summarises this view: a CWM is not a single monolithic predictor, but a sequence of modelling commitments that moves from perception to representation (Rung 1), from representation to causal structure (Rung 1-2), and from structure to intervention (Rung 2), utility, and counterfactual reasoning (Rung 3).
W
e give particular attention to the state-abstraction pipeline at time $t$
\vspace{-0.6cm}

\begin{equation}\label{eq:mappings}
    \rvx_t \xlongrightarrow{\phi} \rvv_t=(\rvv_t^i)_{i\in O_t} \xlongrightarrow{\psi_{O_t}} \rvr_t,\\[-0.15cm]
\end{equation}
where $\phi$ maps the observed random variable $\rvx_t$ to entity-indexed latent random variables $\rvv_t$, with $\rvv_t^i$ for each entity $i$ in the current entity set $O_t$, and $\psi_{O_t}$ assembles the complete structured relational state $\rvr_t$. Its diagonal blocks retain entity attributes and its off-diagonal blocks encode interactions.
The first map $\phi$ raises a \emph{representation-learning question}: under what assumptions can raw observations identify the entities and attributes that should enter the state representation? The second map $\psi_{O_t}$ raises a \emph{structure-learning question}: which attributes and inter-entity relations should be represented as variables, and how can samples from those variables support causal discovery?
These steps are coupled to, but distinct from, transition learning, policy or action modelling, utility specification, and the structural assumptions needed to use action-conditioned transitions to predict the effects of interventions~\citep{bareinboim2022on,wang2022causal_dynamics}.
Additionally, an action model endows CWMs with the capability to evaluate goal oriented actions $\rva_t$ and predict their effects on the structured relational state $\rvr_{t+1}$, as well as a decoder $\hat{f}:\gR_{O_{t+1}}\to\gX$ translating $\rvr_{t+1}$ to the next observation $\rvx_{t+1}$.

Concretely, this work makes two contributions. First, we formalise a CWM as a Markov decision process that links observations, latent states, actions, transition distributions, and utility (Section~\ref{sec:cwm}). Second, we argue that guarantees for CWMs should be understood component-wise, each imposing different assumptions and admitting different admissible equivalences, so that the right notion of identifiability for CWMs is the one that preserves the downstream reasoning task and the corresponding notion of control 
(Section~\ref{sec:wm_identifiability}).
\vspace{-0.4cm}
\section{Causal World Models}
\label{sec:cwm}
\vspace{-0.3cm}
Causal mechanisms govern the dynamics of the world, motivating a formal framework that can capture such mechanisms for a given environment.
Our view of WMs integrates this property by introducing a structured formalisation that progresses from observations to \emph{latent representations (capturing entity-level abstractions)},  extends it to their \emph{interactions}, modelling this as a causal discovery task. 
To ground the discussion, we use a classic example, introducing additional context as needed.
\vspace{-0.1cm}
\begin{example}
    Let's consider a robot learning to act in a tabletop environment containing a red block, a blue block, and a goal region. 
    The robot receives images $\rvx_t \in \gX$ and executes actions $\rva_t \in \gA$, such as pushing the red block left or moving the gripper towards the goal. 
    This is a single modelling problem, but it can be analysed at different levels of abstraction that link the raw pixels to choosing the optimal policy. 
    Here, $\rvx_t$ denotes the raw observation available to the robot, such as an image of the scene. 
The tuple $\rvv_t$ captures entities such as the red block, blue block, gripper, and goal region; each $\rvv^i_t$ encodes attributes such as position, colour, shape, velocity, and contact state. The diagonal block $\rvr_t^{i,i}$ is exactly $\rvv_t^i$, while an off-diagonal block such as $\rvr_t^{i,j}$ can encode the red block leaning on the blue block
.
\end{example}
\vspace{-0.3cm}
We now formalise the notion of relational variables:
\begin{definition}[Relational Variables]
\label{def:relational_variables}
Let $\gV_O=\prod_{i\in O}\gV^i$ be the entity-indexed latent state space for an entity set $O$, and let $\rvv_t=(\rvv_t^i)_{i\in O}\in\gV_O$. For each ordered pair $i,j\in O$, let $M_{ij}$ be a possibly empty component index set; for $i=j$, $M_{ii}$ indexes all coordinates or attribute blocks of $\rvv_t^i$. Define
\[
\rvr_{m,t}^{i,j}:=
\begin{cases}
\rvv_{m,t}^{i}, & i=j,\\
g_m^{i,j}(\rvv_t^i,\rvv_t^j), & i\neq j.
\end{cases}
\]
The corresponding pair-specific node set is
\[
\nodeSet_r^{i,j}=\{\rvr_m^{i,j}:m\in M_{ij}\}.
\]
Thus $\rvr_t^{i,i}\equiv\rvv_t^i$. With
\[
\mathcal I_r(O)=\{\alpha=(i,j,m):i,j\in O,\ m\in M_{ij}\},
\]
write $\rvr_{\alpha,t}\equiv\rvr_{m,t}^{i,j}$. The structured relational state is $\rvr_t=(\rvr_{\alpha,t})_{\alpha\in\mathcal I_r(O)}\in\gR_O$, and $\psi_O$ in Eq.~\ref{eq:mappings} denotes the assembly of these diagonal and off-diagonal blocks. Directed relations may occupy distinct $(i,j)$ and $(j,i)$ blocks; symmetry is imposed only for relation types declared symmetric.
\end{definition}
\begin{definition}[Formal state description]
\label{def:formal_state}
At time $t$, the state is $\rvs_t=(\rvx_t,\rvr_t)\in\gX\times\gR_{O_t}$, where $O_t$ is the current entity set and $\rvr_t=(\rvr_{\alpha,t})_{\alpha\in\mathcal I_r(O_t)}$ is the structured relational state from Definition~\ref{def:relational_variables}.
\end{definition}
\vspace{-0.3cm}
\noindent Relational variables provide the state description on which a CWM operates. They turn entity-level latent representations into variables whose mechanisms can be learned, intervened on, and evaluated against task goals. 
We can therefore define a CWM as the decision model that couples this relational variable to observations, actions, transitions, and utility.
\vspace{-0.1cm}
\begin{definition}[Causal World Models (CWM)]
\label{def:WM}
A CWM is a tuple
\(
W =
\bigl(
\gX,\gA,\{\gR_O\}_{O},\mathbb{P},\gU
\bigr),
\)
where $\gX \subseteq \bigcup_i \mathbb{R}^{d_i}$ is the observation space, $\gA$ is the action space, $\gR_O$ is the relational latent-state space for entity set $O$, and $\gU:\gA\times\gS\rightarrow\mathbb{R}$ is a utility function over actions and states.
For time $t$, let $\rvx_t,\rvx_{t+1}\in\gX$, $\rva_t\in\gA$, $\rvr_t\in\gR_{O_t}$, and $\rvr_{t+1}\in\gR_{O_{t+1}}$.
Under the Markovian state assumption, the observed one-step decision distribution is obtained by marginalising the latent relational variables,
\vspace{-0.1cm}
\[
\mathbb{P}(\rvx_t,\rvx_{t+1},\rva_t)
=
\int_{\rvr_t,\rvr_{t+1}}
\mathbb{P}(\rvx_t,\rvx_{t+1},\rvr_t,\rvr_{t+1},\rva_t),\\[-0.1cm]
\]
The action-conditioned observation-level transition then factorises as
\vspace{-0.1cm}
\[
\begin{aligned}
\mathbb{P}_{xx'}(a)
&=
\mathbb{P}(\rvx_{t+1}=x' \mid \rvx_t=x,\rva_t=a) \\
&=\!\!
\int_{\rvr_t,\rvr_{t+1}}\!\!\!\!\!\!\!\!\!\!\!\!\!
\mathbb{P}(\rvx_{t+1}\!\mid\!\rvr_{t+1})
\mathbb{P}(\rvr_{t+1}\!\mid\!\rvr_t,\rva_t=a)
\mathbb{P}(\rvr_t\!\mid\!\rvx_t).
\end{aligned}
\]
\end{definition}
\begin{remark}
\label{rem:action_transition}
A CWM can be read as a Markov decision process over the formal state $\rvs_t$ in Definition~\ref{def:formal_state}. The utility $\gU$ evaluates a considered action and state with respect to the agent's goal; in particular, one selects actions satisfying $\gU(a,\rvs_t)\geq \gU(a',\rvs_t)$ for alternatives $a'$.
The factorisation in Definition~\ref{def:WM} has the following components:   
\vspace{-0.2cm}
    \begin{enumerate}
        \item $\mathbb{P}(\rvx_{t+1} \mid \rvr_{t+1}) \rightarrow$ Prediction model, generates next observation given $\rvr_{t+1}$, following the structure in $\gG_x$,  
\vspace{-0.1cm}
        \item $\mathbb{P}(\rvr_{t+1}\mid\rvr_t) \rightarrow$ Transition model, with $\mathbb{P}(\rva_t\mid\rvr_t)$ as the action model and $\mathbb{P}(\rvr_{t+1}\mid\rvr_t,\rva_t)$ as the action-conditioned transition. For discrete actions,
        $\mathbb{P}(\rvr_{t+1}\mid\rvr_t)=\sum_{a\in\gA}\mathbb{P}(\rvr_{t+1}\mid\rvr_t,a)\mathbb{P}(a\mid\rvr_t)$.
        Under consistency and positivity, when actions are randomised or, more generally, there is no unmeasured action--outcome confounding given $\rvr_t$~\citep{hernan2020causal}, this conditional identifies the controlled transition $\mathbb{P}(\rvr_{t+1}\mid\rvr_t;\operatorname{do}(\rva_t=a))$, where $\operatorname{do}(\rva_t=a)$ denotes externally setting the action to $a$~\citep{pearl_2009}; otherwise it is only a predictive conditional,
\vspace{-0.1cm}
        \item $\mathbb{P}(\rvr_t \mid \rvx_t) \rightarrow$ Inference model, first infers the entity-centric representation $\rvv_t$ and then assembles the structured relational state $\rvr_t$, where $\mathbb{P}(\rvr_t \mid \rvx_t) = \int_{\rvv_t} \mathbb{P}(\rvr_t \mid \rvv_t) \mathbb{P}(\rvv_t \mid \rvx_t)$, while following the structure $\gG_r$.
    \end{enumerate}
\end{remark}
\vspace{-0.3cm}
\noindent  The definition is deliberately abstract: it does not prescribe a particular estimator, but identifies the probabilistic components that will later be associated with the literature on representation learning, causal discovery, and causal decision-making.
The remark separates a CWM into components that must be learned or specified: an inference model from observations to relational variables, a transition or intervention model over relational variables, an action model, a prediction model back to observations, and a utility function. 
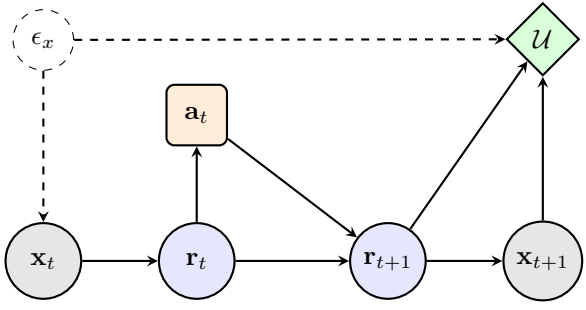
\begin{figure}[!t] 
    \centering
    \begin{tikzpicture}
        [
            node distance=1.0cm and 1.0cm,
            >=stealth,
            every node/.style={font=\normalsize},
            latent/.style={
                circle, draw=black, thick,
                fill=blue!10,
                minimum size=1cm
            },
            obs/.style={
                circle, draw=black, thick,
                fill=gray!20,
                minimum size=1cm
            },
            decision/.style={
                rectangle, draw=black, thick,
                fill=orange!15,
                minimum width=0.8cm,
                minimum height=0.8cm,
                rounded corners=3pt
            },
            utility/.style={
                diamond, draw=black, thick,
                fill=green!15,
                minimum width=0.8cm,
                minimum height=0.8cm,
                aspect=1
            },
            noise/.style={
                circle, draw=black, dashed,
                minimum size=0.8cm
            },
        ]
        
        \node[obs] (xt) {$\rvx_t$};
        \node[latent, right=of xt] (vt) {$\rvr_t$};
        \node[decision, above=1cm of vt] (at) {$\rva_t$};
        \node[latent, right=1.5cm of vt] (vtp1) {$\rvr_{t+1}$};
        \node[obs, right=of vtp1] (xtp1) {$\rvx_{t+1}$};
        
        \node[utility, above=1.9cm of xtp1] (u) {$\gU$};
        \node[noise, above=2cm of xt] (epsx) {$\epsilon_x$};
        
        \draw[->, thick] (xt) -- (vt);
        \draw[->, thick] (vt) -- (vtp1);
        \draw[->, thick] (vt) -- (at);
        \draw[->, thick] (at) -- (vtp1);
        \draw[->, thick] (vtp1) -- (xtp1);
        
        \draw[->, thick] (xtp1) -- (u);
        \draw[->, thick] (vtp1) -- (u);
        
        \draw[->, thick, dashed] (epsx) -- (xt);
        \draw[->, thick, dashed] (epsx) -- (u);
        
    \end{tikzpicture}
    \caption{\textbf{Causal influence diagram for the CWM factorisation}. Circles denote random variables: grey circles are observations and blue circles are relational variables. The orange rectangle is the decision node, and the green diamond is the utility node. Solid arrows denote the dependencies in Definition~\ref{def:WM}: $\rvx_t\mapsto\rvr_t$ is inference, $(\rvr_t,\rva_t)\mapsto\rvr_{t+1}$ is transition, $\rvr_{t+1}\mapsto\rvx_{t+1}$ is prediction
    . Dashed arrows indicate possible latent confounders.\\[-0.8cm]}

    \label{fig:WM-CID}
\end{figure}
\vspace{-0.5cm}

\noindent Figure~\ref{fig:WM-CID} instantiates Definition~\ref{def:WM} as a causal influence diagram, a graphical model that combines causal structure with decision and utility nodes~\citep{everitt2021agent}. In our setting, the observation $\rvx_t$ is encoded into relational variables $\rvr_t$, the action $\rva_t$ selects or modifies the transition mechanism leading to $\rvr_{t+1}$, and the next observation $\rvx_{t+1}$ is decoded from the updated relational state. The utility $\gU$ evaluates the outcome relative to the agent's goal, such as moving the red block into the goal region without undesired collisions. The diagram makes explicit that a WM should not only predict the next observation, but also support goal-oriented reasoning about which actions are useful for a task.

\noindent With this interpretation in place, we can state the \emph{causal sufficiency} assumption behind Definition~\ref{def:WM}, as shown in Figure~\ref{fig:WM-CID}. A WM can support causal reasoning only if the variables that jointly drive prediction and utility are available to the model, either as recorded components of the observation or as latent factors inferred from it. If a relevant common cause is missing from both levels, it may act as an unobserved confounder, possibly distorting causal inferences. This is the standard causal sufficiency assumption in causal discovery~\citep{spirtes2000causation}, which we make here to focus on the identifiability of the latent structure rather than the additional challenges of learning with \emph{unobserved confounders}.
Under this assumption, the CWM definition organises the modelling problem component-wise. If $\rvx_t$ has already been decomposed into the causal variables of interest, the task may be causal discovery over $\gG_x$ together with action-conditioned transition learning. For unstructured observations such as raw pixels, we do not posit a pixel-level causal DAG; instead, causal representation learning recovers $\rvv_t$ and the structured state $\rvr_t$, and causal discovery targets the causal graph $\gG_r$.
\section{CWM Identifiability}
\label{sec:wm_identifiability}
Identifiability asks which aspects of a CWM are uniquely determined by a data regime. Exact recovery is one possible target, but latent-variable models more commonly admit transformations that leave the observable distribution unchanged~\citep{khemakhem2020ice,kivva2022identifiability,von2021self}. Causal graphs and interventional quantities introduce different observational ambiguities~\citep{verma1990equivalence,pearl_2009,bareinboim2022on}; hence identifiability must be stated for a particular component and downstream use.
Intuitively, the question is whether two candidate CWMs compatible with the same evidence can still disagree on a subset of their components; the definitions below distinguish exact agreement from disagreements declared admissible.
\begin{definition}[Strong identifiability]
Let $\mathfrak{W}$ be a class of candidate WMs, and let 
$\mathbb{P}_{\mathfrak{D}}(W)$ denote the observable distribution induced by $W \in \mathfrak{W}$ under a data regime $\mathfrak{D}$. The class $\mathfrak{W}$ is strongly identifiable from $\mathfrak{D}$ if, for any $W_1,W_2 \in \mathfrak{W}$,
\(
\mathbb{P}_{\mathfrak{D}}(W_1)=\mathbb{P}_{\mathfrak{D}}(W_2) \Rightarrow
W_1 = W_2 .
\)
\label{def: swm_identifiability}
\end{definition}
\vspace{-0.2cm}
Strong identifiability is rarely the right target for CWMs. Entity permutations and invertible latent reparameterisations, including rescaling, can leave the learnt mechanisms unchanged. For example, two CWMs may place the red and blue blocks in opposite slots; after applying the same permutation to both relational indices and the entities affected by each action, they may induce identical predictions and decisions despite having unequal parameterisations. Requiring $W_1=W_2$ would distinguish such semantically equivalent models: \emph{equivalence} expresses this admissible variability.
\begin{definition}[Identifiability up to equivalence]
Let $\sim$ be an equivalence relation over $\mathfrak W$, where $W_1 \sim W_2$ means that the two models differ only by transformations that preserve the relevant world-model semantics. The class $\mathfrak W$ is identifiable up to $\sim$ from $\mathfrak D$ if
\(
\mathbb{P}_{\mathfrak{D}}(W_1)=\mathbb{P}_{\mathfrak{D}}(W_2)
 \Rightarrow
W_1 \sim W_2.
\)
\label{def: wwm_identifiability}
\end{definition}
\vspace{-0.6cm}
For the CWM (Def.~\ref{def:WM}), Table~\ref{tab:cwm_component_identifiability} makes the equivalence $\sim$ of Definition~\ref{def: wwm_identifiability} concrete across its representation, distributions, structure and utility, combining known recovery results with bespoke interface conditions. It distinguishes \emph{representation equivalences} for $\rvv_t$, $\{\rvv_t^i\}$, $\rvv_{B,t}^i$, and $\rvr_t^{i,j}$ from \emph{compatibility conditions} on the CWM interfaces. These equivalences jointly induce a representation alignment $T$, with $\tilde{\rvr}=T(\rvr)$ and, on formal states (Def.~\ref{def:formal_state}), $T(\rvx,\rvr)=(\rvx,T(\rvr))$. We write $\overset{d}{=}$ for equality in distribution.

\begin{table*}[!t]
\centering
\begin{tabular}{p{0.25\linewidth}p{0.7\linewidth}}
\toprule
\textbf{CWM Component}
& \textbf{Admissible Equivalence or Compatibility Condition} \\
\midrule

$\rvv_t$: Latent state  
& Invertible affine equivalence $\tilde{\rvv}_t=\mH\rvv_t+\rvc$; under stronger mixture assumptions this reduces to permutation, coordinate-wise scaling, and translation~\citep{kivva2022identifiability}. \\

$\{\rvv_t^i\}_{i\in O_t}$: Entity blocks
& Permutation of entity/slot indices: $\tilde{\rvv}_t^i=\rvv_t^{\pi(i)}$~\citep{locatello2020object}. \\

$\rvv_{B,t}^i$: Entity-attribute block
& Within-entity affine identifiability~\citep{kori2024identifiable}; for a declared attribute partition, the block-preserving specialisation is $\tilde{\rvv}_{B,t}^i=\mH_B\rvv_{B,t}^{\pi(i)}+\rvc_B^i$. \\

$\rvr_t^{i,j}$, $i\neq j$: Inter-entity relational block
& Under our alignment of compositional blocks, following~\citep{kori2025unifying}, require entity permutation with element-wise affine compatibility:
$\tilde{\rvr}_t^{i,j}=\mD_{ij}\rvr_t^{\pi(i),\pi(j)}+\rvc_{ij}$. Diagonal blocks inherit the equivalence of $\rvv_t^i$. \\

$\psi_O$: Relational assembly map
& Equivariance under a common permutation of entity labels in both indices; order invariance is required only for relation types declared symmetric. \\

$p_{\hat{f}_{\sharp}}(\rvx_{t+1}\mid\rvr_{t+1})$: Prediction model  
& Weak injectivity of the decoder push-forward \citep{kivva2022identifiability} and local inverse consistency. With $E_O:=\psi_O\circ\phi$, require $E_O\approx\hat f^{-1}$ on the relevant support: $E_O(\hat f(\rvr))\approx\rvr$ and $\hat f(E_O(\rvx))\approx\rvx$ on the data manifold. \\

$\mathbb{P}(\rvr_{t+1}\mid\rvr_t)$: Transition model  
& Equivariance in distribution under state alignment:
$\tilde{\rvr}_{t+1}\mid T(\rvr)
\overset{d}{=}T(\rvr_{t+1})\mid\rvr$. \\

$\gG_r$: Relational-variable causal graph  
& Markov equivalence class (MEC) over the relational-variable DAG: same skeleton and same v-structures after admissible variable alignment \citep{verma1990equivalence}. \\

$\mathbb P(\rvr_{t+1}\mid\rvr_t,\rva_t)$: Action-conditioned transition
& Actions and their affected components correspond under $T$, and
$\tilde{\rvr}_{t+1}\mid T(\rvr),a
\overset{d}{=}T(\rvr_{t+1})\mid\rvr,a$.
Causal interpretation additionally requires consistency, positivity, and either randomised actions or no unmeasured action--outcome confounding given $\rvr_t$~\citep{hernan2020causal}; cf.~Remark~\ref{rem:action_transition}. \\

$\gU$: Utility model
& Utility preservation under transformation $T$: $\tilde{\gU}(a,T(\rvs))=\gU(a,\rvs)$. \\
\bottomrule
\end{tabular}
\caption{Component-wise equivalences and compatibility conditions for causal world-model identifiability.\\[-0.35cm]}
\label{tab:cwm_component_identifiability}
\end{table*}

The representation conditions form a hierarchy of semantic commitments. The coordinates of $\rvv_t$ need not have fixed names, whereas $\{\rvv_t^i\}$ must preserve entity correspondence. Since \citet{kori2024identifiable} identify each whole entity vector $\rvv_t^i$, interpreting $\rvv_{B,t}^i$ as a named attribute block additionally requires the transformation to preserve that partition. For $\rvr_t^{i,j}$, the same entity correspondence must act on both indices and retain the diagonal aliases. The compositional construction of \citet{kori2025unifying} motivates combining such blocks, without establishing their causal identifiability.
\vspace{-0.15cm}

Compatibility concerns whether the aligned representation can be used coherently by the rest of the CWM. Equivariance of $\psi_O$ makes entity relabelling yield the same assembled relational state, while local inverse consistency of $E_O=\psi_O\circ\phi$ and $\hat f$ prevents the encoder--decoder interface from discarding relevant state information. The passive and action-conditioned dynamics and $\gU$ must respect the same $T$ so that aligned states produce aligned rollouts and values. The causal graph $\gG_r$ has a different status: its MEC is itself a graph equivalence~\citep{verma1990equivalence}, not an interface condition, and need not preserve action effects. Matched actions must affect components carried to one another by $T$: entity-directed actions follow the entity permutation, whereas relation-directed actions follow both relational indices. Action-conditioned transitions support causal planning only under the stated identification conditions (cf.~Remark~\ref{rem:action_transition}). Thus, equivalences specify what may change within individual components; compatibility conditions ensure that these changes propagate coherently through the CWM, yielding an admissible equivalence of the full model.
Component-wise guarantees compose into a policy guarantee only when these interfaces are compatible. In particular, if the encoder and decoder satisfy the stated local inverse-consistency condition, the action-conditioned transition commutes with $T$ in distribution, and actions, their affected components, and utilities are preserved under the same alignment, then corresponding policies induce aligned trajectory distributions and the same expected cumulative utility, for the same horizon and discount. Their utility-maximising policies therefore correspond under $T$, as in policy-preserving MDP abstractions~\citep{li2006stateabstraction}. If transition effects, intervention targets, or utilities fail to commute with the same alignment, the per-component equivalences do not compose into a control guarantee.
In summary, the relevant target is the weakest component-wise equivalence whose compatible interfaces preserve the intended prediction, intervention, or decision objective. 
This separates what can be identified from data from what must be supplied as structural assumptions or expert knowledge.
\vspace{-0.4cm}
\section{Conclusion}
\vspace{-0.3cm}
This paper argues that CWMs should be understood as structured decision models rather than as monolithic predictors. 
We formalised a CWM as a model linking observations, representations and structure, and used this formalisation to connect representation learning, causal discovery, and model-based decision-making. 
The central contribution is a component-wise view, where each part of the model may be identifiable only up to admissible equivalence that enables the downstream reasoning. 
This decomposition clarifies what can be learned from data, what must be supplied by interventions or domain knowledge, and when predictive models can support causal reasoning. 
Future work should turn this perspective into a rigorous formalisation, followed by practical algorithms and empirical testing, while extending it to partial observability and latent confounding.

\begin{acknowledgements}
Russo was funded by the ERC under the ERC-POC programme (grant number 101189053) and Kori was support by EPSRC post doctoral prize fellowship.
\end{acknowledgements}

\bibliographystyle{plainnat}
\renewcommand{\bibsection}{\subsubsection*{References}}
\bibliography{bibfile}


\end{document}

%% file: math_notations.tex
\usepackage{amsmath,amsfonts,bm}

\def\eqref#1{equation~\ref{#1}}

\def\1{\bm{1}}

\def\rva{{\mathbf{a}}}

\def\rvc{{\mathbf{c}}}

\def\rvr{{\mathbf{r}}}
\def\rvs{{\mathbf{s}}}

\def\rvv{{\mathbf{v}}}

\def\rvx{{\mathbf{x}}}

\def\mD{{\bm{D}}}

\def\mH{{\bm{H}}}

\DeclareMathAlphabet{\mathsfit}{\encodingdefault}{\sfdefault}{m}{sl}
\SetMathAlphabet{\mathsfit}{bold}{\encodingdefault}{\sfdefault}{bx}{n}

\def\gA{{\mathcal{A}}}

\def\gG{{\mathcal{G}}}

\def\gR{{\mathcal{R}}}
\def\gS{{\mathcal{S}}}

\def\gU{{\mathcal{U}}}
\def\gV{{\mathcal{V}}}

\def\gX{{\mathcal{X}}}



%% file: bibfile.bib
@article{ZANGA2022survey,
  author       = {Alessio Zanga and
                  Elif Ozkirimli and
                  Fabio Stella},
  title        = {A Survey on Causal Discovery: Theory and Practice},
  journal      = {Int. J. Approx. Reason.},
  volume       = {151},
  pages        = {101--129},
  year         = {2022},
  opturl          = {https://doi.org/10.1016/j.ijar.2022.09.004},
  bibsource    = {dblp computer science bibliography, https://dblp.org}
}

@article{ge2026review,
  title={A review of causal decision making},
  author={Ge, Lin and Cai, Hengrui and Wan, Runzhe and Xu, Yang and Song, Rui},
  journal={Journal of Artificial Intelligence Research},
  volume={85},
  year={2026}
}

@inproceedings{verma1990equivalence,
  author = {Verma, Thomas and Pearl, Judea},
  title = {Equivalence and synthesis of causal models},
  year = {1990},
  optisbn = {0444892648},
  optpublisher = {Elsevier Science Inc.},
  optaddress = {USA},
  optbooktitle = {Proceedings of the Sixth Annual Conference on Uncertainty in Artificial Intelligence},
  booktitle = {Proc. of UAI},
  pages = {255–270},
  numpages = {16},
  optseries = {UAI '90},
  opturl = {https://ftp.cs.ucla.edu/tech-report/1991-reports/910020.pdf},
}

@book{spirtes2000causation,
  author       = {Peter Spirtes and
                  Clark Glymour and
                  Richard Scheines},
  title        = {Causation, Prediction, and Search, Second Edition},
  series       = {Adaptive computation and machine learning},
  publisher    = {{MIT} Press},
  year         = {2000},
  optisbn         = {978-0-262-19440-2},
  bibsource    = {dblp computer science bibliography, https://dblp.org},
  opturl          = {https://www.cs.cmu.edu/afs/andrew/scs/cs/15-381/archive/OldFiles/lib/cvsub/.g/group/sdss/.g/group2/g/scottd/fullbook.pdf}
}

@book{pearl2018bookofwhy,
author = {Pearl, Judea and Mackenzie, Dana},
title = {The Book of Why: The New Science of Cause and Effect},
year = {2018},
optisbn = {046509760X},
publisher = {Basic Books, Inc.},
address = {USA},
edition = {1st}
}

@article{scholkopf2021toward,
  author       = {Bernhard Sch{\"{o}}lkopf and
                  Francesco Locatello and
                  Stefan Bauer and
                  Nan Rosemary Ke and
                  Nal Kalchbrenner and
                  Anirudh Goyal and
                  Yoshua Bengio},
  title        = {Toward Causal Representation Learning},
  journal      = {Proc. {IEEE}},
  volume       = {109},
  number       = {5},
  pages        = {612--634},
  year         = {2021},
  opturl          = {https://doi.org/10.1109/JPROC.2021.3058954},
  bibsource    = {dblp computer science bibliography, https://dblp.org}
}

@article{ha2018world,
  title={World models},
  author={Ha, David and Schmidhuber, J{\"u}rgen},
  journal={arXiv:1803.10122},
  volume={2},
  number={3},
  year={2018}
}

@article{ding2025understanding,
  title={Understanding world or predicting future? a comprehensive survey of world models},
  author={Ding, Jingtao and Zhang, Yunke and Shang, Yu and Zhang, Yuheng and Zong, Zefang and Feng, Jie and Yuan, Yuan and Su, Hongyuan and Li, Nian and Sukiennik, Nicholas and others},
  journal={ACM Computing Surveys},
  volume={58},
  number={3},
  pages={1--38},
  year={2025},
  publisher={ACM New York, NY}
}

@article{taniguchi2023world,
  title={World models and predictive coding for cognitive and developmental robotics: frontiers and challenges},
  author={Taniguchi, Tadahiro and Murata, Shingo and Suzuki, Masahiro and Ognibene, Dimitri and Lanillos, Pablo and Ugur, Emre and Jamone, Lorenzo and Nakamura, Tomoaki and Ciria, Alejandra and Lara, Bruno and others},
  journal={Advanced Robotics},
  volume={37},
  optnumber={13},
  pages={780--806},
  year={2023},
  publisher={Taylor \& Francis}
}

@article{matsuo2022deep,
  title={Deep learning, reinforcement learning, and world models},
  author={Matsuo, Yutaka and LeCun, Yann and Sahani, Maneesh and Precup, Doina and Silver, David and Sugiyama, Masashi and Uchibe, Eiji and Morimoto, Jun},
  journal={Neural Networks},
  volume={152},
  pages={267--275},
  year={2022},
  publisher={Elsevier}
}

@article{lecun2022path,
  title={A path towards autonomous machine intelligence version 0.9. 2, 2022-06-27},
  author={LeCun, Yann},
  journal={Open Review},
  optvolume={62},
  optnumber={1},
  optpages={1--62},
  year={2022}
}

@InProceedings{guan2023leveraging,
  title={Leveraging pre-trained large language models to construct and utilize world models for model-based task planning},
  author={Guan, Lin and Valmeekam, Karthik and Sreedharan, Sarath and Kambhampati, Subbarao},
  optbooktitle={Advances in Neural Information Processing Systems},
  booktitle={Proc. of NeurIPS},
  volume={36},
  pages={79081--79094},
  year={2023}
}

@article{hafner2025dreamer,
  title={Mastering diverse control tasks through world models},
  author={Hafner, Danijar and Pasukonis, Jurgis and Ba, Jimmy and Lillicrap, Timothy},
  journal={Nature},
  volume={640},
  pages={647--653},
  year={2025},
  optdoi={10.1038/s41586-025-08744-2},
  opturl={https://www.nature.com/articles/s41586-025-08744-2}
}

@article{xu2026resdreamer,
  title={Self-supervised Hierarchical Visual Reasoning with World Model},
  author={Xu, Yuanfei and Liu, Lin and Zhou, Wengang and Feng, Mingxiao and Li, Houqiang},
  journal={arXiv:2605.17537},
  year={2026}
}

@article{klindt2026lejeppa,
  title={When Does {LeJEPA} Learn a World Model?},
  author={Klindt, David and LeCun, Yann and Balestriero, Randall},
  year={2026},
  journal={arXiv:2605.26379},
  eprint={2605.26379},
  archivePrefix={arXiv},
  primaryClass={stat.ML},
  optdoi={10.48550/arXiv.2605.26379},
  opturl={https://arxiv.org/abs/2605.26379}
}

@inproceedings{
richens2025general,
title={General agents need world models},
author={Jonathan Richens and Tom Everitt and David Abel},
optbooktitel={Forty-second International Conference on Machine Learning},
booktitle={Proc. of ICML},
year={2025},
opturl={https://openreview.net/forum?id=dlIoumNiXt}
}

@inproceedings{
richens2024robust,
title={Robust agents learn causal world models},
author={Jonathan Richens and Tom Everitt},
optbooktitel={The Twelfth International Conference on Learning Representations},
booktitle={Proc. of ICLR},
year={2024},
opturl={https://openreview.net/forum?id=pOoKI3ouv1}
}

@inproceedings{ceriscioli2025agents,
  title={Agents Robust to Distribution Shifts Learn Causal World Models Even Under Mediation},
  author={Ceriscioli, Matteo and Mohan, Karthika},
  optbooktitel={The Thirty-ninth Annual Conference on Neural Information Processing Systems},
  booktitle={Proc. of NeurIPS},
  year={2025},
  opturl={https://neurips.cc/virtual/2025/loc/san-diego/poster/118687},
}

@inproceedings{everitt2021agent,
  title={Agent Incentives: A Causal Perspective},
  author={Everitt, Tom and Carey, Ryan and Langlois, Eric D. and Ortega, Pedro A. and Legg, Shane},
  optbooktitel={Proceedings of the AAAI Conference on Artificial Intelligence},
  booktitle={Proc. of AAAI},
  volume={35},
  optnumber={13},
  pages={11487--11495},
  year={2021},
  opturl={https://ojs.aaai.org/index.php/AAAI/article/view/17368}
}

@inproceedings{wang2022causal_dynamics,
  title={Causal Dynamics Learning for Task-Independent State Abstraction},
  author={Wang, Zizhao and Xiao, Xuesu and Xu, Zifan and Zhu, Yuke and Stone, Peter},
  optbooktitel={Proceedings of the 39th International Conference on Machine Learning},
  booktitle={Proc. of ICML},
  optseries={Proceedings of Machine Learning Research},
  volume={162},
  pages={23151--23180},
  year={2022},
  optpublisher={PMLR},
  opturl={https://proceedings.mlr.press/v162/wang22ae.html}
}

@inproceedings{wang2024causal_bisimulation,
  title={Building Minimal and Reusable Causal State Abstractions for Reinforcement Learning},
  author={Wang, Zizhao and Wang, Caroline and Xiao, Xuesu and Zhu, Yuke and Stone, Peter},
  optbooktitel={AAAI Conference on Artificial Intelligence},
  booktitle={Proc. of AAAI},
  year={2024},
  opturl={https://www.cs.utexas.edu/~pstone/Papers/bib2html/b2hd-cbm-wang-aaai24.html}
}

@inproceedings{li2006stateabstraction,
  title={Towards a Unified Theory of State Abstraction for {MDP}s},
  author={Li, Lihong and Walsh, Thomas J. and Littman, Michael L.},
  booktitle={Proc. of the International Symposium on Artificial Intelligence and Mathematics},
  year={2006},
  opturl={https://anytime.cs.umass.edu/aimath06/proceedings/P21.pdf}
}

@book{hernan2020causal,
  title={Causal Inference: What If},
  author={Hernan, Miguel A. and Robins, James M.},
  publisher={Chapman \& Hall/CRC},
  year={2020},
  opturl={https://www.hsph.harvard.edu/miguel-hernan/causal-inference-book/}
}

@inproceedings{kori2024identifiable,
  title={Identifiable object centric representations via probabilistic slot attention},
  author={Kori, Avinash and Locatello, Francesco and Toni, Francesca and Glocker, Ben and Ribeiro, Fabio De Sousa},
  optbooktitel={Part of Advances in Neural Information Processing Systems 37 (NeurIPS 2024)},
  booktitle={Proc. of NeurIPS},
  year={2024}
}

@inproceedings{
kori2025unifying,
title={Unifying Causal and Object-centric Representation Learning allows Causal Composition},
author={Avinash Kori and Ben Glocker and Bernhard Sch{\"o}lkopf and Francesco Locatello},
optbooktitel={ICLR 2025 Workshop on Deep Generative Model in Machine Learning: Theory, Principle and Efficacy},
booktitle={Proc. of ICLR Workshop on Deep Generative Model in Machine Learning: Theory, Principle and Efficacy},
year={2025},
opturl={https://openreview.net/forum?id=KBW866rTgL}
}

@InProceedings{locatello2020object,
  title={Object-centric learning with slot attention},
  author={Locatello, Francesco and Weissenborn, Dirk and Unterthiner, Thomas and Mahendran, Aravindh and Heigold, Georg and Uszkoreit, Jakob and Dosovitskiy, Alexey and Kipf, Thomas},
  optbooktitle={Advances in Neural Information Processing Systems},
  booktitle={Proc. of NeurIPS},
  volume={33},
  pages={11525--11538},
  year={2020}
}

@article{khemakhem2020ice,
  title={Ice-beem: Identifiable conditional energy-based deep models based on nonlinear ica},
  author={Khemakhem, Ilyes and Monti, Ricardo and Kingma, Diederik and Hyvarinen, Aapo},
  journal={Advances in Neural Information Processing Systems},
  volume={33},
  pages={12768--12778},
  year={2020}
}

@InProceedings{kivva2022identifiability,
  title={Identifiability of deep generative models without auxiliary information},
  author={Kivva, Bohdan and Rajendran, Goutham and Ravikumar, Pradeep and Aragam, Bryon},
  optbooktitle={Advances in Neural Information Processing Systems},
  booktitle={Proc. of NeurIPS},
  volume={35},
  pages={15687--15701},
  year={2022}
}

@article{von2021self,
  title={Self-supervised learning with data augmentations provably isolates content from style},
  author={Von K{\"u}gelgen, Julius and Sharma, Yash and Gresele, Luigi and Brendel, Wieland and Sch{\"o}lkopf, Bernhard and Besserve, Michel and Locatello, Francesco},
  journal={Advances in neural information processing systems},
  volume={34},
  pages={16451--16467},
  year={2021}
}

@book{pearl_2009, place={Cambridge}, edition={2}, title={Causality}, DOI={10.1017/CBO9780511803161}, publisher={Cambridge University Press}, author={Pearl, Judea}, year={2009}}

@article{bengio2013generalized,
  title={Generalized denoising auto-encoders as generative models},
  author={Bengio, Yoshua and Yao, Li and Alain, Guillaume and Vincent, Pascal},
  journal={Advances in neural information processing systems},
  volume={26},
  year={2013}
}

@inbook{bareinboim2022on,
author = {Bareinboim, Elias and Correa, Juan D. and Ibeling, Duligur and Icard, Thomas},
title = {On Pearl’s Hierarchy and the Foundations of Causal Inference},
year = {2022},
optisbn = {9781450395861},
publisher = {Association for Computing Machinery},
optaddress = {New York, NY, USA},
edition = {1},
opturl = {https://doi.org/10.1145/3501714.3501743},
booktitle = {Probabilistic and Causal Inference: The Works of Judea Pearl},
pages = {507–556},
numpages = {50}
}
